%% file: example.tex
\documentclass{article}
\usepackage{graphics} 
\usepackage{epsfig} 
\usepackage{multirow}
\usepackage{booktabs} 
\usepackage{array}
\usepackage{amsmath} 
\usepackage{amssymb}  
\usepackage{lipsum}
\usepackage{pifont}
\usepackage{graphicx}
\usepackage{subcaption}
\usepackage{wrapfig}
\usepackage{amsmath}
\usepackage{amssymb}
\usepackage{graphicx}
\usepackage{booktabs}
\usepackage{multirow}
\usepackage{lipsum}
\usepackage{pifont}
\usepackage{graphicx}
\usepackage{subcaption}
\usepackage[table]{xcolor}
\definecolor{lightblue}{RGB}{220,235,250}

\usepackage{xspace}

\newcommand{\mname}{\textsl{CDP}}

\usepackage[final]{corl_2025} 

\title{ CDP: Towards Robust Autoregressive Visuomotor Policy Learning via Causal Diffusion }

\author{
  Jiahua Ma$^{*}$ \\
  Sun Yat-sen University \\
  \texttt{majiahua99@gmail.com} \\
  \And
  Yiran Qin$^{*\ddagger}$ \\
  CUHK(SZ); Oxford \\
  \texttt{yiranqin@link.cuhk.edu.cn} \\
  \And
  Yixiong Li \\
  Sun Yat-sen University \\
  \texttt{liyx357@mail2.sysu.edu.cn} \\
  \And
  Xuanqi Liao \\
  Sun Yat-sen University\\
  \texttt{liaoxq7@mail2.sysu.edu.cn} \\
  \And
  Yulan Guo \\
  Sun Yat-sen University\\
  \texttt{guoyulan@sysu.edu.cn} \\
  \And
  Ruimao Zhang$^{\dagger}$ \\
  Sun Yat-sen University\\
  \texttt{zhangrm27@mail.sysu.edu.cn} \\
}

\begin{document}
\maketitle

\renewcommand*{\thefootnote}{}
\footnotetext[1]{$^*$Equal contribution. $^\ddagger$Project Leader. $^\dagger$Corresponding author. }


\begin{abstract}
Diffusion Policy (\textsl{DP}) enables robots to learn complex behaviors by imitating expert demonstrations through action diffusion. However, in practical applications, hardware limitations often degrade data quality, while real-time constraints restrict model inference to instantaneous state and scene observations. These limitations seriously reduce the efficacy of learning from expert demonstrations, resulting in failures in object localization, grasp planning, and long-horizon task execution.
To address these challenges, we propose Causal Diffusion Policy (\mname{}), a novel transformer-based diffusion model that enhances action prediction by conditioning on historical action sequences, thereby enabling more coherent and context-aware visuomotor policy learning.
To further mitigate the computational cost associated with autoregressive inference, a caching mechanism is also introduced to store attention key-value pairs from previous timesteps, substantially reducing redundant computations during execution.
Extensive experiments in both simulated and real-world environments, spanning diverse 2D and 3D manipulation tasks, demonstrate that \mname{} uniquely leverages historical action sequences to achieve significantly higher accuracy than existing methods. Moreover, even when faced with degraded input observation quality, \mname{} maintains remarkable precision by reasoning through temporal continuity, which highlights its practical robustness for robotic control under realistic, imperfect conditions. Project page: \url{https://gaavama.github.io/CDP/}
\end{abstract}



\section{Introduction}
\label{sec:introduction}
Robotic manipulation tasks—such as grasping, placing, and assembling objects—are notoriously difficult to program explicitly due to their high complexity and variability. 
Recently, imitation learning (\textsl{IL}) offers a promising alternative, allowing robots to acquire manipulation skills by mimicking expert demonstrations~\cite{shridhar2023perceiver, wang2023mimicplay, ze2023gnfactor, peng2020learning, agarwal2023dexterous, haldar2023teach}. 
By learning from example trajectories, \textsl{IL} eliminates the need for manual behavior design, making it particularly effective in contact-rich or dynamic environments where reward functions or controllers are difficult to define.
\begin{figure*}[t]
    \centering
    \setlength{\fboxrule}{0pt}
    \framebox{{\includegraphics[width=\linewidth]{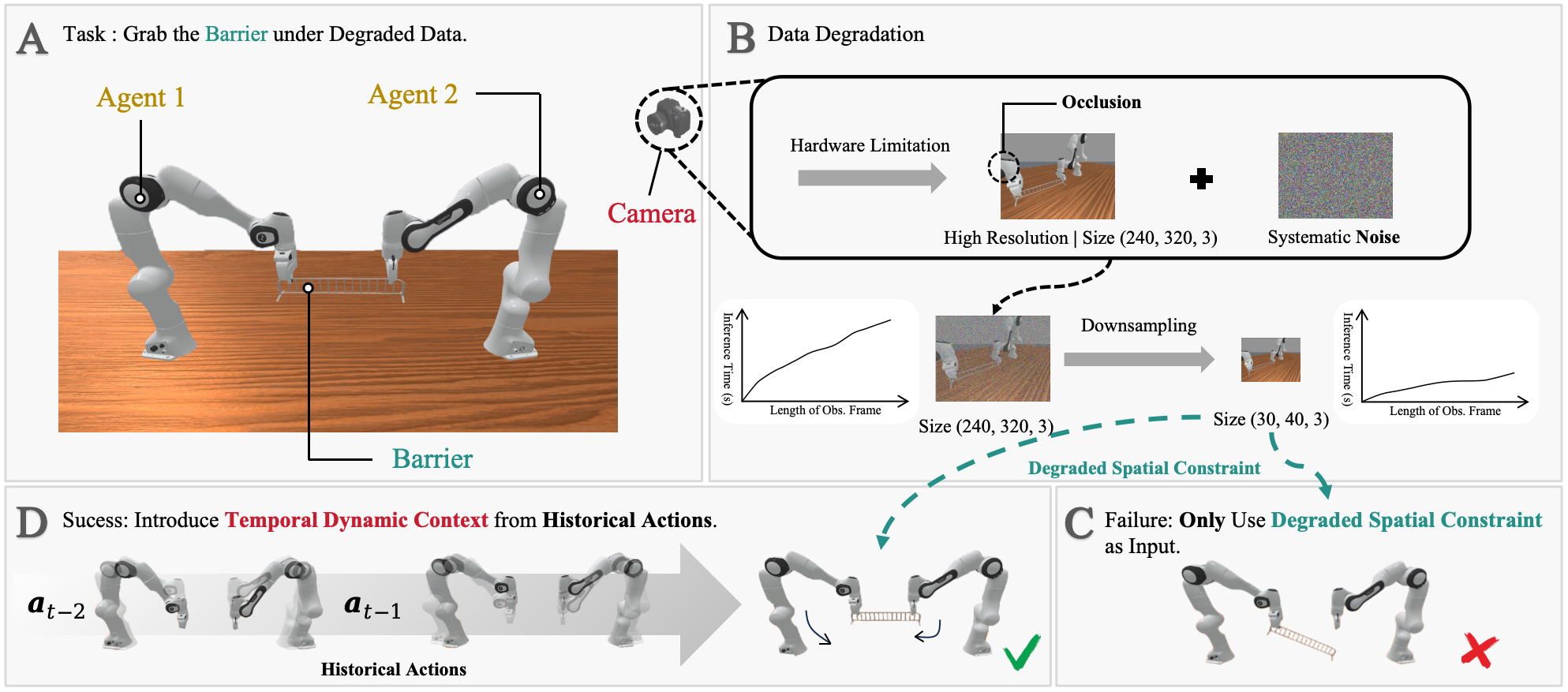}}}
    \caption{ \textbf{Causal Diffusion Policy}: a transformer-based diffusion model that enhances action prediction by conditioning on historical action sequences. \textbf{A}: When performing the task of “grabbing the barrier” in practice, \textbf{B}: the quality of observations is degraded by factors such as sensor noise, occlusions, and hardware limitations. In fact, this degraded but high-dimensional observation data not only fails to provide sufficient spatial constraint information for policy planning but also slows down the planning speed. \textbf{C}: In this case, the robot is unable to perform accurate manipulation. \textbf{D}: In this paper, we address historical action sequences to introduce temporally rich context as a supplement, which enables more robust policy generation.}
    \label{fig:intro}
\end{figure*}

As a powerful \textsl{IL} framework, Diffusion Policy~(\textsl{DP})~\cite{chi2023diffusion} treats action generation as a denoising diffusion process, which enable smooth and multimodal trajectory prediction, paving the way for more versatile and efficient robot learning system.
However, \textsl{DP} employs a naive behavior cloning approach to learn the specified tasks, which models actions independently and fails to account for the sequential structure of decision making.
This often leads to \emph{distributional shift}, where small prediction errors accumulate and push the robot into unseen or unstable states. 

Furthermore, robot actions are continuous and temporally correlated—properties that are difficult to capture under observations common in physical deployments.
And in real-world scenarios shown in Fig.~\ref{fig:intro}, sensor noise, occlusions, and hardware limitations degrade observation quality, while real-time inference constraints restrict access to temporally rich context.
As a result, \textsl{DP} often fails at critical subtasks such as object localization, grasp planning, and long-horizon task execution.

To address these issues, we propose \textbf{Causal Diffusion Policy (\mname{})}, a novel transformer-based diffusion framework that explicitly conditions action prediction on historical action sequences. 
Rather than relying solely on spatial constraints in instantaneous observations, \mname{} incorporates temporal continuity through an autoregressive causal transformer, enabling the policy to reason over prior actions and their evolving contexts. 
This design improves coherence, stability, and robustness, especially in challenging conditions where single-frame observations may be unreliable.

To further improve computational efficiency during inference, we introduce a \textbf{caching mechanism} that stores and reuses attention key-value pairs computed in earlier timesteps. 
This avoids redundant computations inherent in standard transformer-based policies and makes the model practical for real-time deployment on physical hardware. 
Together, causal modeling and inference caching allow \mname{} to generate temporally consistent and high-quality actions with reduced latency.
Our contributions can be summarized as follows:
\begin{enumerate}
    \item We propose \mname{}, a novel transformer-based diffusion framework for robotic manipulation that conditions action prediction on historical action sequences, enabling context-aware, temporally coherent visuomotor policy learning.
    \item To enable efficient real-time inference, we introduce a cache sharing mechanism that stores and reuses attention key-value pairs from prior timesteps, substantially reducing computational overhead in autoregressive action generation.
    \item Through extensive evaluations on diverse 2D and 3D manipulation tasks in both simulation and real-world settings, we show that \mname{} achieves superior accuracy and robustness over existing methods, especially under degraded observation conditions.
\end{enumerate}

\section{Related Work}
\label{sec:related_work}

\subsection{Diffusion Model in Robotic Manipulation}
Diffusion Model~\cite{ho2020denoising, song2020denoising} provides an effective means for robots to acquire human-like skills by emulating expert demonstrations~\cite{janner2022planning, luo2024potential, carvalho2023motion, saha2024edmp,zhou2024minedreamer,qin2025navigatediff,  huang2023diffusion}. Recent advancements~\cite{pearce2023imitating,ha2023scaling, xian2023chaineddiffuser,li2024crossway, wang2024poco, chen2024don, sridhar2023memory, zhao2024aloha, chi2024universal, kang2025viki} in this domain have introduced innovative approaches to model visuomotor policies. Diffusion Policy~\cite{chi2023diffusion} models a visuomotor policy as a conditional denoising diffusion process, i.e. utilizing the robot’s observations as the condition to refine noisy trajectories into coherent action sequences. Based on this, 3D Diffusion Policy~\cite{Ze2024DP3} incorporates simple point-cloud representations, thereby enriching the robot’s perception of the environment. 
%
Despite these advancements, these holistic approaches that generate complete action sequences in a single pass face inherent limitations, such as potential error propagation and challenges in handling long-range dependencies within the action sequences. To address these issues, recent research has increasingly focused on token-wise incremental generation for robot policy learning. This paradigm shift aims to improve the flexibility and robustness of action generation by breaking down the process into smaller, manageable steps.
Notable examples include ICRT~\cite{fu2024context}, ARP~\cite{zhang2025autoregressive}, and CARP~\cite{gong2024carp}.
In this work, we build upon these foundational advancements by proposing a novel transformer-based causal generation model. By incorporating temporal context, our model is better equipped to capture the dynamics of the environment and generate smoother, more coherent action sequences. 

\subsection{Causal Generation in Diffusion Model}
Diffusion models have utilized uniform noise levels across all tokens during training. While this approach simplifies the training process, it may compromise the model's ability to capture complex temporal dynamics~\cite{chen2023pixart, peebles2023scalable,qin2024worldsimbench, rombach2022high}. To address this limitation, Diffusion Forcing~\cite{chen2024diffusion} introduced a novel training strategy for sequence diffusion models, where noise levels are independently varied for each frame. This method significantly enhances the model's capacity to generate more realistic and coherent sequences, as demonstrated by both theoretical analysis and empirical results. Building on this foundation, CausVid~\cite{yin2024slow} further extended Diffusion Forcing by integrating it into a causal transformer architecture, resulting in an autoregressive video foundation model. This adaptation effectively combines the strengths of both Diffusion Forcing and transformer-based architectures, thereby achieving improved temporal coherence and consistency in video generation~\cite{guo2023animatediff, lu2023vdt, ma2024latte, ren2024consisti2v,yu2025position,yu2025survey,yu2025gamefactory,yu2025cam}. Ca2-VDM~\cite{gao2024ca2} incorporates optimized causal mechanisms and efficient cache management techniques. These innovations reduce computational redundancy by reusing precomputed conditional frames and minimize storage costs through cache sharing across denoising steps, thereby enhancing real-time performance.
%

\section{Method}
\label{sec:method}
\subsection{Causal Action Generation}
\label{sec:training}

We begin by outlining the training phase of our proposed model, followed by a detailed description of the Causal Action Generation Module (Fig.~\ref{fig:training}~(a)). Based on this, we introduce the Causal Temporal Attention Mechanism, a pivotal component of the module. This mechanism ensures that each target action can access all historical actions, thereby capturing its temporal dynamics context and enabling the generation of more accurate results. In addition, to enhance the module's robustness against error-prone historical actions and mitigate the risk of action generation failure due to the accumulation of prediction errors during inference phase, we further integrate the Historical Actions Enhancement. This module significantly improves reliability of the action generation process.
\begin{figure*}[t]
    \centering
    \setlength{\fboxrule}{0pt}
    \framebox{{\includegraphics[width=\linewidth]{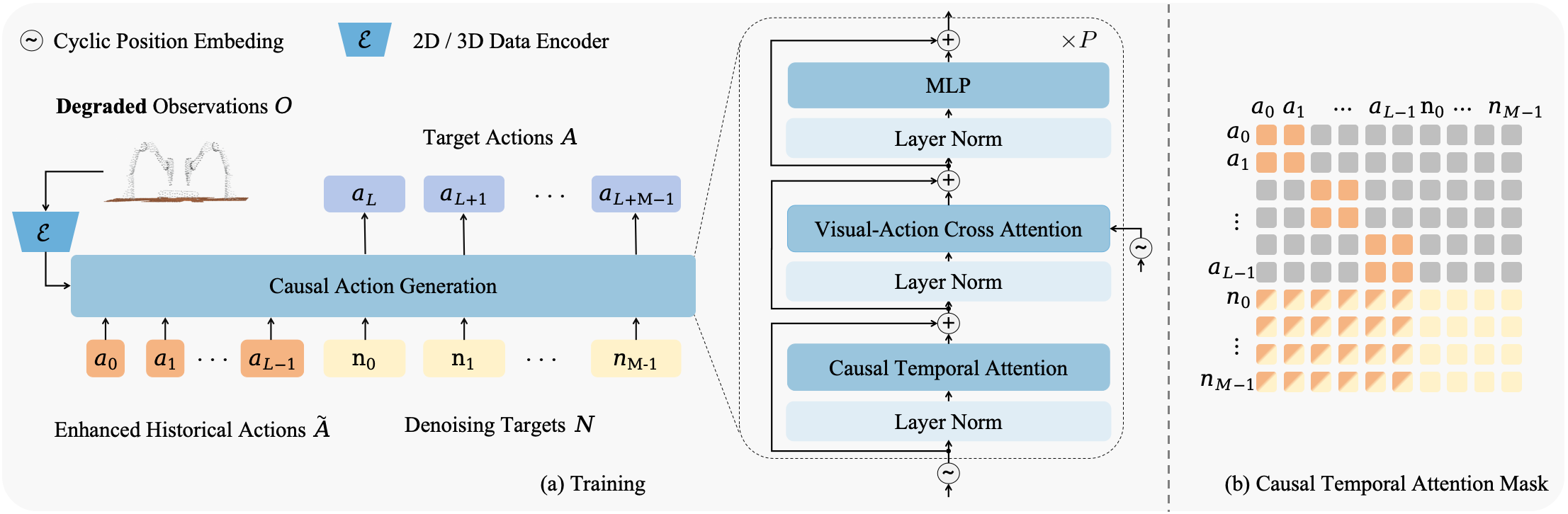}}}
    \caption{ \textbf{Causal Action Generation of Our \mname{}.} (a) During training, the Historical Actions \( \tilde{\textbf{A}} \) are combined with the Denoising Targets \( \textbf{N} \). This combined input is then fed into the Causal Action Generation module, which contains \( P \) blocks, for denoising. 
    The Target Actions \( \textbf{A} \) are used for training supervision. Before denoising, \( \tilde{\textbf{A}} \) is perturbed by a small-scale noise, which helps to reduce the accumulation of action prediction errors during inference. 
    (b) The Causal Temporal Attention Mask ensures each Denoising Target can access all Historical Actions.}
    \label{fig:training}
\end{figure*}

\textbf{Training}.
During training, the sampled action sequence with the length of $L+M$ is divided into two parts, i.e. the Historical Actions  
$\tilde{\textbf{A}} = \{a_k\}_{k=0}^{L-1}$
with the length of $L$, and the Target Actions 
$\textbf{A} = \{a_k\}_{k=L}^{L+M-1}$
with the length of $M$. The former serves as the causal condition, which will be perturbed and then, fed into the Causal Action Generation Module together with the Denoising Targets $\textbf{N} = \{n_k \sim \mathcal{N}(0, 1)\}_{k=0}^{M-1}$ and ultimately generate the predicted Target Actions. The objective function for the training of this module is designed to minimize the L2 distance between the predicted Target Actions and its corresponding ground truth,
\begin{equation}
  \min \mathbb{E}_{\tilde{\textbf{A}}, \textbf{A}, \textbf{N}} \| (D_\theta([\tilde{\textbf{A}}, \textbf{N}]) - \textbf{A}) \|^2_2   
\end{equation}
where \( [\cdot, \cdot] \) denotes the concatenation operation along the temporal axis and \( D_\theta \) denotes the whole denoising process, based on our Causal Action Generation Module with the parameter of \( \theta \).

\textbf{Historical Actions Re-Denoising}. In autoregressive prediction tasks, errors in historical actions can cause significant deviations that accumulate over time, potentially leading to the failure of robotic manipulation.
To address this issue, we perturb the Historical Actions \(\tilde{\textbf{A}}\) by introducing small-scale noise and then, re-denoise them during training.
\begin{equation}
  \tilde{\textbf{A}}^\texttt{perturb} =  \tilde{\textbf{A}}~+~\textbf{N}_\sigma
\end{equation}
where $\tilde{\textbf{A}}^\texttt{perturb}$ denotes the perturbed historical actions and $\textbf{N}_\sigma \sim \mathcal{N}(0, \sigma^2)$ denotes the small-scale noise sequence with the variance of $0 < \sigma < 1$.
By introducing $\textbf{N}_\sigma$, we create a more robust training environment that simulates the occasions where Historical Actions \(\tilde{\textbf{A}}\) may be imperfect or noisy, forcing the model to focus on the coarse-grained temporal dynamic context from the Historical Actions \(\tilde{\textbf{A}}\), instead of depending on their concrete values. These coarse-grained temporal dynamics serve as a complement to the degraded observations \(\textbf{O}\), guiding the subsequent denoising process.

\textbf{Causal Action Generation Module}.
As depicted in Fig.~\ref{fig:training}~(a), this module is composed of $P$ blocks. 
Within each block, we first inject the temporal dynamic context from the Historical Actions \(\tilde{\textbf{A}}\) into the Denoising Targets \(\textbf{N}\) through Causal Temporal Attention ($\texttt{CTA}$), generating intermediate features \( \mathbf{N}_{t} \) .
Following this, Visual-Action Cross Attention ($\texttt{VACA}$) is utilized to incorporate spatial constraints from the degraded observation $\textbf{O}$ into these \( \mathbf{N}_{t} \), generating intermediate features \( \mathbf{N}_{ts} \). These \( \mathbf{N}_{ts} \) are then further processed by a Multi-layer Perceptron ($\texttt{MLP}$). It is worth noting that  we place the denoising timestep embedding $t$ in the $\texttt{VACA}$ layer, rather than the $\texttt{CTA}$ layer in order to ensure the cached features of the Historical Actions \(\tilde{\textbf{A}}\) can be correctly shared during the whole denoising process (Sec.~\ref{sec:inference}). The entire process within each block can be described as follows,
\begin{align}
\mathbf{N}_{t} &= \texttt{LN}~(\texttt{CTA}~(\tilde{\textbf{A}},~~\mathbf{N})) \\
\mathbf{N}_{ts} &= \texttt{LN}~(\texttt{VACA}~(\texttt{Enc}~(\mathbf{O}), ~~\mathbf{N}_{t}) \\
\mathbf{N}_{o}&= \texttt{LN}~(\texttt{MLP}~(\mathbf{N}_{ts}))
\end{align}
where \( \mathbf{N}_{o} \) represents the output features of this block.
$\texttt{LN}$ denotes the Layer Normalization operation, and $\texttt{Enc}$ denotes the 2D or 3D data encoder.

\textbf{Causal Temporal Attention ($\texttt{CTA}$)}. To ensure the causality of action prediction while maintaining compatibility with the inference phase, our Causal Temporal Attention must meet the following criteria: 
1) During the autoregressive inference process, new actions are generated based on their historical actions. This means that we only need to ensure that the Denoising Targets \(\textbf{N}\) can attend to the Historical Actions \(\tilde{\textbf{A}}\) during the attention calculation, rather than enforcing causality among the Historical Actions \(\tilde{\textbf{A}}\) themselves. 
2) The autoregressive inference process continually discards invalid historical actions, which are too early and too distant from the Target Actions \(\textbf{A}\) temporally. This implies that during the attention calculation, we need to further decouple the Historical Actions \(\tilde{\textbf{A}}\). Specifically, we chunk the Historical Actions \(\tilde{\textbf{A}}\) with a fixed size, ensuring that all actions within a chunk are fully visible to one another, while actions from different chunks cannot attend to each other. 
3) Following~\cite{chi2023diffusion}, we increase the redundancy in the length of the Denoising Targets \(\textbf{N}\) to enable the model to anticipate future context while maintaining action coherence throughout the entire inference process. Ultimately, only the first chunk-size actions are utilized as the final predicted Target Actions.
To meet the aforementioned constraints, we customize the Causal Temporal Attention Mask, as illustrated in Fig.~\ref{fig:training}~(b). For the Historical Action \(a_0\), the corresponding mask  
\raisebox{-0.25\height}{\includegraphics[height=0.35cm]{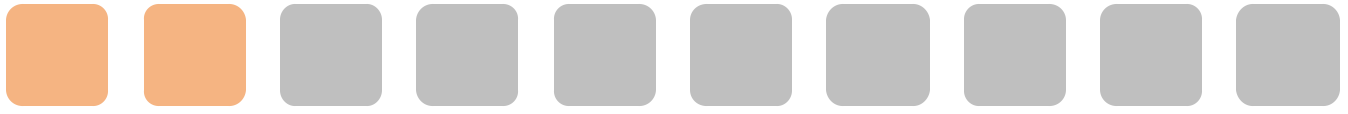}}  
indicates that the attention computation only considers actions within the same chunk  
\raisebox{-0.25\height}{\includegraphics[height=0.35cm]{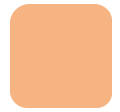}},  
excluding actions from other chunks  
\raisebox{-0.25\height}{\includegraphics[height=0.35cm]{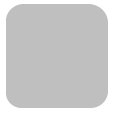}}.  
Furthermore, all Denoising Targets \(n_i\) are treated as being in a single chunk with a larger chunk size than that of the Historical Actions. Their corresponding mask  
\raisebox{-0.24\height}{\includegraphics[height=0.35cm]{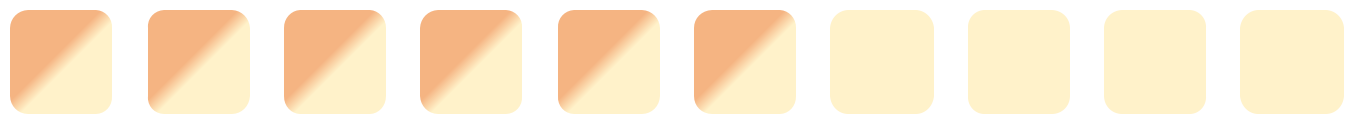}}  
indicates that their attention calculation considers both the Historical Actions  
\raisebox{-0.25\height}{\includegraphics[height=0.35cm]{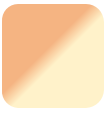}}  
and the Denoising Targets themselves  
\raisebox{-0.25\height}{\includegraphics[height=0.35cm]{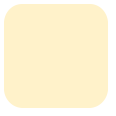}}.

\begin{figure*}[t]
    \centering
    \setlength{\fboxrule}{0pt}
    \framebox{{\includegraphics[width=\linewidth]{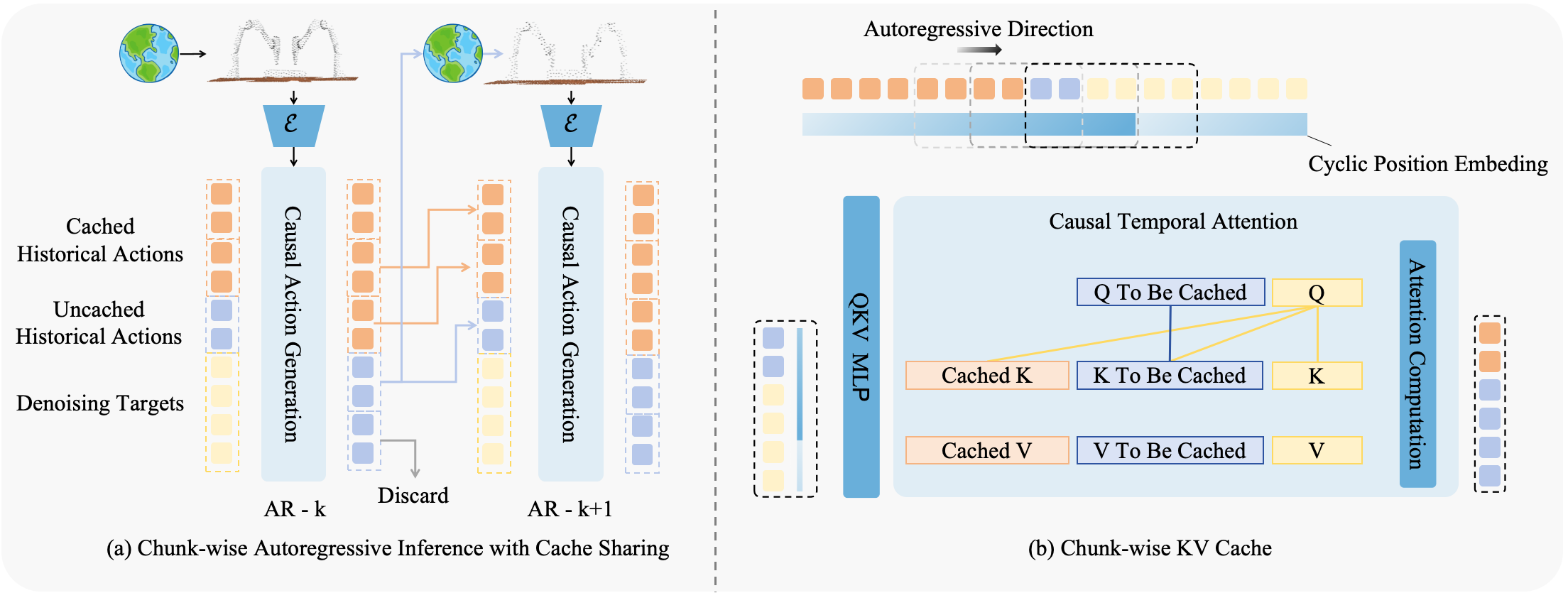}}}
    \caption{\textbf{Chunk-wise Autoregressive Inference of Our \mname{}.} (a) The orange and purple blocks denote actions whose Key and Value representations have been cached and not cached until the current step, respectively. The yellow block denotes the Gaussian noises. During the \(\texttt{AR-k}\) step, we perform denoising while simultaneously computing and storing the Key and Value representations for the Uncached Historical Actions. After denoising, the Target Actions generated in the \(\texttt{AR-k}\) are applied to the environment and serve as the Uncached Historical Actions in the \(\texttt{AR-k+1}\) step, which are then used to generate future Target Actions.
    (b) In current autoregressive inference step, the Key and Value representations for the Cached Historical Actions can be directly used by the Attention Computation. But for the Uncached Historical Actions and Denoising Targets, we need utilize a \(\texttt{QKV\_MLP}\) layer to extract their Query, Key and Value representations. During Attention Computation, the Uncached Historical Actions are restricted to considering only actions within its own chunk (indicated by the purple line), while the Denoising Targets has access to the entire action sequence (indicated by the yellow line).}
    \label{fig:inference}
\end{figure*}

\subsection{Chunk-wise Autoregressive Inference with Cache Sharing}
\label{sec:inference}
In this section, we first provide an overview of our proposed Chunk-wise Autoregressive Inference process (Fig.~\ref{fig:inference}~(a)), which is equipped with a Cache Sharing mechanism (Fig.~\ref{fig:inference}~(b)). Since the entire target action sequence is predicted in an autoregressive manner, its total length remains indeterminate at the outset. To this end, we introduce Cyclic Temporal Embedding to distinguish the temporal order of the whole action sequence (Fig.~\ref{fig:inference}~(b)).

\textbf{Chunk-wise Autoregressive Inference.} 
%
During the \(k\)-th autoregressive (\(\texttt{AR-k}\)) step, we perform denoising using
\begin{equation}
\mathbf{A}^k \sim p_\theta(\mathbf{A}^k | \mathbf{N}^k, \tilde{\textbf{A}}^k)
\end{equation}
where \(\mathbf{N}^k\), \(\tilde{\textbf{A}}^k\), and \(\mathbf{A}^k\) denote the Denoising Targets, Historical Actions, and Target Actions at the \(\texttt{AR-k}\) step, respectively.
Within the causal generation framework, features computation proceeds in a unidirectional manner. Specifically, \(\textbf{N}^k\) is denoised conditioned on \(\tilde{\textbf{A}}^k\), and the cache for \(\tilde{\textbf{A}}^k\) can be precomputed in prior autoregressive steps without regard for \(\mathbf{N}^k\). After denoising, the Target Actions \(\textbf{A}^{k}\) generated in the \(\texttt{AR-k}\) step are applied to the environment and serve as the partial of Historical Actions in the \(\texttt{AR-k+1}\) step, as illustrated in Fig.~\ref{fig:inference}~(a). In the \(\texttt{AR-k+1}\) step, we update the historical actions in a window-sliding manner (Fig.~\ref{fig:inference}~(b)). This involves discarding the invalid part of \(\tilde{\textbf{A}}^k\) and incorporating the Target Actions \(\mathbf{A}^k\) to form the Historical Actions \(\tilde{\textbf{A}}^{k+1}\) for the \(\texttt{AR-k+1}\) step. These Historical Actions \(\tilde{\textbf{A}}^{k+1}\) are then fed into the Causal Action Generation Module to generate the Target Actions \(\textbf{A}^{k+1}\) in the \(\texttt{AR-k+1}\) step.
  
\textbf{Cache Sharing.} In the \(\texttt{AR-k}\) step, the Historical Actions \(\tilde{\textbf{A}}^k\) of length \(L\) can be divided into two parts: Cached Historical Actions \(\tilde{\textbf{A}}^k_{0:l-1}\) and Uncached Historical Actions \(\tilde{\textbf{A}}^k_{l:L-1}\), based on whether their Key and Value representations have been cached up to this step. Here, \(l\) denotes the length of the Cached Historical Actions. 
During denoising, the Key and Value representations for the Cached Historical Actions \(\tilde{\textbf{A}}^k_{0:l-1}\) have been precomputed in prior autoregressive steps and can be directly used in attention computation. We denote them as \(\mathbf{K}_{0:l-1}^{(0)}\) and \(\mathbf{V}_{0:l-1}^{(0)}\), where \((\cdot)\) indicates the denoising timestep and \((0)\) means these Key and Value representations correspond to the final Target Actions generated in prior steps.
For the Uncached Historical Actions \(\tilde{\textbf{A}}^k_{l:L-1}\) and Denoising Targets \(\textbf{N}^k\), we utilize a \(\texttt{QKV\_MLP}\) layer to extract their Query, Key, and Value representations,
\begin{align}
\{ \mathbf{Q}_{l:L-1}^{(0)}, \mathbf{K}_{l:L-1}^{(0)}, \mathbf{V}_{l:L-1}^{(0)}\} &= \texttt{QKV\_MLP}~(\tilde{\textbf{A}}^k_{l:L-1}) \\
\{ \mathbf{Q}_{L:L+M-1}^{(t)}, \mathbf{K}_{L:L+M-1}^{(t)}, \mathbf{V}_{L:L+M-1}^{(t)}\} &= \texttt{QKV\_MLP}~(\textbf{N}^k)
\end{align}
where \((t)\) indicates that those Query, Key and Value representations correspond to the Denoising Targets at timestep \(t\). Then, \(\mathbf{Q}_{l:L-1}^{(0)}, \mathbf{K}_{l:L-1}^{(0)}, \mathbf{V}_{l:L-1}^{(0)}\) are cached and will be shared across all denoising timesteps. Concatenating all the above representations forms the whole Query, Key, and Value representations (i.e. \(\mathbf{Q}(k, t)\), \(\mathbf{K}(k, t)\), and \(\mathbf{V}(k, t)\)) at denoising timestep \(t\) in the \(\texttt{AR-k}\) step.
\begin{align}
\mathbf{Q}(k, t) &= \texttt{Concat}(\mathbf{Q}_{l:L-1}^{(0)},~ \mathbf{Q}_{L:L+M-1}^{(t)}) \\
\mathbf{K}(k, t) &= \texttt{Concat}(\mathbf{K}_{0:l-1}^{(0)},~ \mathbf{K}_{l:L-1}^{(0)},~ \mathbf{K}_{L:L+M-1}^{(t)}) \\
\mathbf{V}(k, t) &= \texttt{Concat}(\mathbf{V}_{0:l-1}^{(0)},~ \mathbf{V}_{l:L-1}^{(0)},~ \mathbf{V}_{L:L+M-1}^{(t)})
\end{align}
where \(\texttt{Concat}\) denotes the concatenation operation along temporal dimension.
Subsequently, the causal temporal attention is computed as:
\begin{equation}
\mathbf{Q}(k, t) \times [\mathbf{K}(k, t) \times \mathbf{V}(k, t) + \tilde{\textbf{M}}]
\end{equation}
where \(\tilde{\textbf{M}}\) denotes the attention mask with dimensions \((L-l+M) \times (L+M)\), mirroring the training attention mask illustrated in Fig.~\ref{fig:training}(b). As shown in Fig.~\ref{fig:inference}(b), this mask ensures that: 1) The attention weights for the Uncached Historical Actions are computed regardless of any other actions. 2) The prediction of Target Actions are based on all Historical Actions.

\textbf{Cyclic Temporal Embedding.} In traditional temporal embedding mechanisms, unique temporal embeddings are typically assigned to each action during the inference phase. However, this method is not effective due to the variable length of action sequences. To tackle this problem, we present the Cyclic Temporal Embedding mechanism. Instead of generating new embeddings, the Denoising Targets utilize temporal embeddings that are cyclically shifted from the beginning of the sequence. During the training process, a random cyclic offset is applied to the temporal embedding sequence of each sample. This randomization enables the model to generalize across various temporal offsets, thereby enhancing its robustness in the inference phase.

\section{Simulation Experiments}
\label{sec:sim_experiment}

\begin{table}[t!]
    \centering
    \renewcommand{\arraystretch}{1.5}
    \caption{Quantitative evaluation of our \mname{} and baseline methods on \textbf{simulation tasks}, highlighting the effectiveness of our approach.  
    }
    \label{tab:sim-performance}
    \resizebox{\textwidth}{!}{
    \begin{tabular}{c c c c c c c c c c c}
        \hline
        \multirow{2}{*}{Method} & \multicolumn{2}{c}{Adroit} & \multicolumn{2}{c}{DexArt} & \multicolumn{4}{c}{MetaWorld} & \multicolumn{2}{c}{RoboFactory} \\ \cmidrule(lr){2-3} \cmidrule(lr){4-5} \cmidrule(lr){6-9} \cmidrule(lr){10-11}
        & hammer & Pen & Laptop & Toilet & Bin Picking & Box Close & Disassemble & Reach & Lift Barrier & Place Food \\ \hline
        \rowcolor{lightblue}
        CDP3~(Ours) & \textbf{100\%} & \textbf{68\%} & \textbf{84\%} & \textbf{68\%} & \textbf{43\%} & \textbf{54\%} & \textbf{83\%} & \textbf{39\%} & \textbf{40\%}  & \textbf{19\%} \\ 
        DP3~\cite{Ze2024DP3} & \textbf{100\%} & 49\% & 81\% & 65\% & 38\% & 42\% & 69\% & 24\% & 35\% & 13\% \\ 
        \hline
        \rowcolor{lightblue}
        CDP~(Ours) & \textbf{100\%} & \textbf{37\%} & \textbf{52\%} & \textbf{38\% } & \textbf{24\%} & \textbf{35\%} & \textbf{51\%} & \textbf{38\%} & 22\% & \textbf{22\%} \\ 
        DP~\cite{chi2023diffusion} & \textbf{100\%} & 29\% & 31\% & 26\% & 21\% & 30\% & 43\% & 27\% & \textbf{29\%} & 17\% \\ \hline
    \end{tabular}
    }
\end{table}

\subsection{Quantitative Results}

The quantitative results of all methods on the simulation benchmarks are detailed in Table~\ref{tab:sim-performance}, which illustrates that our \mname{} consistently outperforms the baseline methods across both 2D and 3D scenarios. Specifically, \mname{} achieves an improvement of about 5\% to 20\% points over the baseline methods across tasks of varying difficulty levels in different benchmarks.
These findings highlight that, with an identical visual encoder, the causal generation paradigm employed by \mname{} significantly enhances the success rate of downstream tasks. This success can be attributed to two key aspects. First, the causal generation approach leverages historical actions by learning its temporal dynamic context to guide the generation of future actions. This temporal context provides valuable information that aids in making more informed and coherent decisions. Second, \mname{} effectively addresses the challenges caused by the degraded observations. In scenarios where the current observation alone cannot provide sufficient information for action generation, \mname{} utilizes the historical action sequence as a supplementary information. This dual reliance on both observations and historical actions ensures robustness in policy planning.

\subsection{Ablation Study}
\label{sec:ablation}
\begin{figure*}[t!]   
  \centering
  \setlength{\fboxrule}{0pt}
  \begin{minipage}[b]{0.49\linewidth}
    \centering
    \framebox{\includegraphics[width=\linewidth]{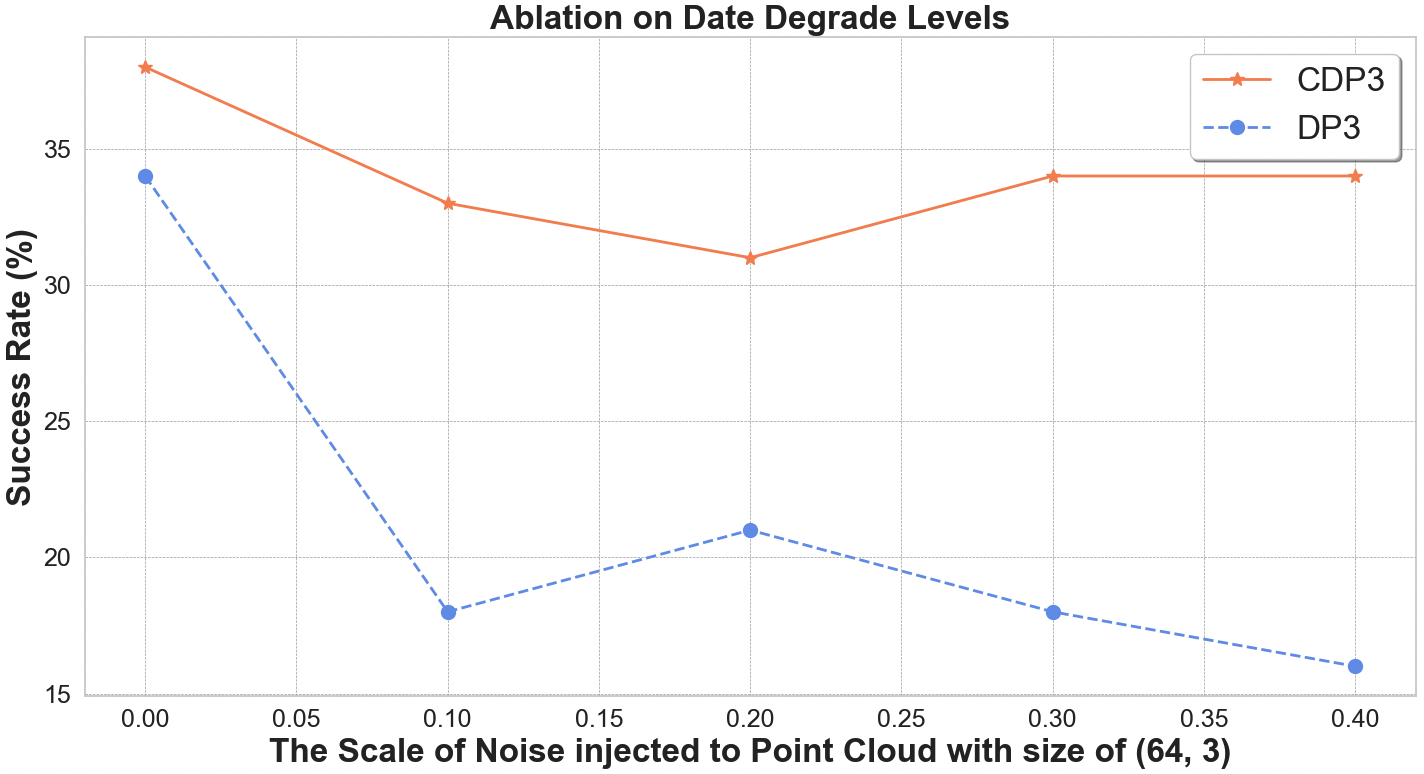}}
    \subcaption{}\label{fig:degraded_level}
  \end{minipage}
  \hfill
  \begin{minipage}[b]{0.49\linewidth}
    \centering
    \framebox{\includegraphics[width=\linewidth]{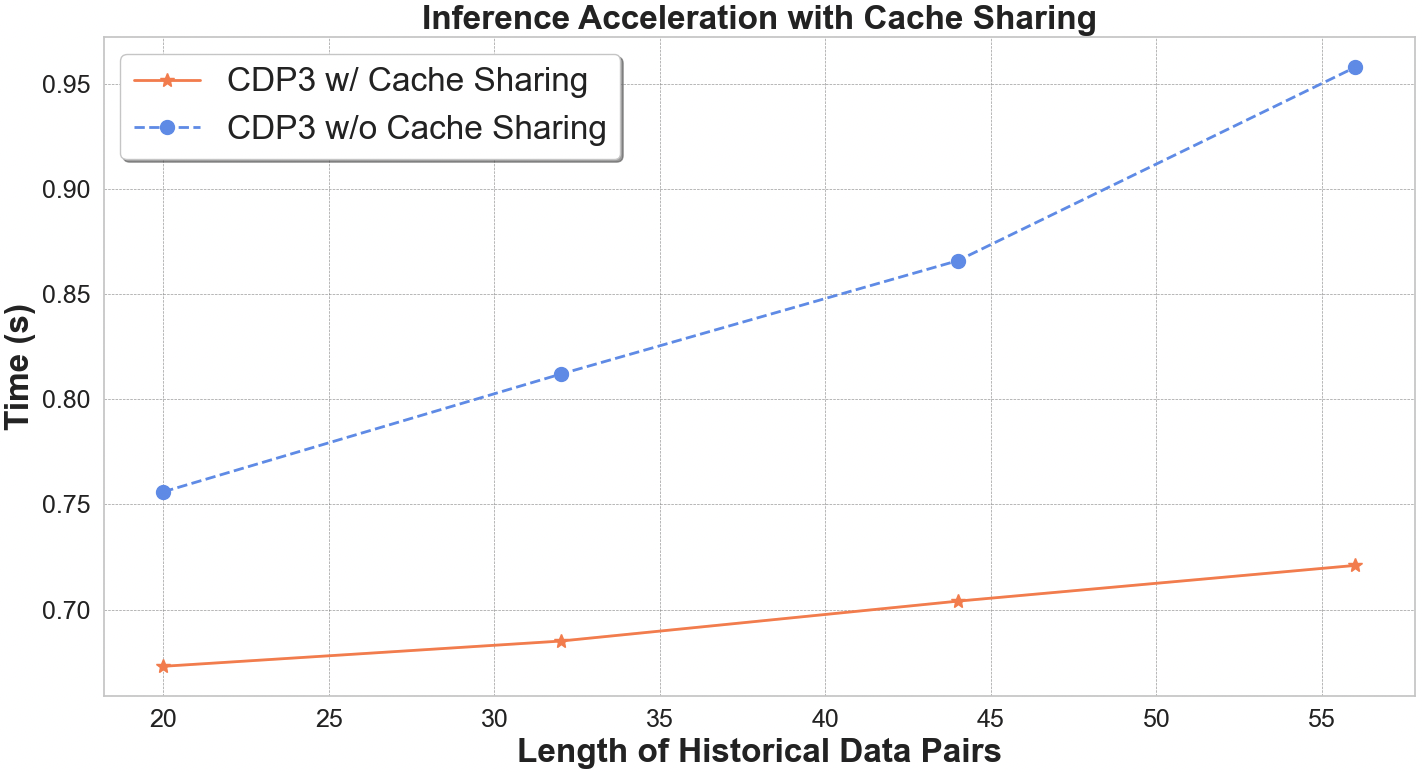}}
    \subcaption{}\label{fig:acceleration_level}
  \end{minipage}
  
  \caption{\textbf{Visualization of the Effectiveness and Efficiency of Our \mname{}.} 
  (a) The performance of DP3 collapses as observation degrades, whereas our \mname{} remains nearly unaffected.
  (b) Our Cache Sharing Mechanism yields inference-time speed-ups that scale approximately linearly with the length of the Historical Actions.}
  \label{fig:training}
\end{figure*}

In this section, we conduct experiments on the \textbf{Lift Barrier} task from RoboFactory~\cite{qin2025robofactory} to rigorously evaluate both the effectiveness and efficiency of our proposed \mname{}.

\noindent \textbf{Robustness to Degraded Observation Data}. 
In this experiment, we introduce varying levels of noise to the low-resolution point clouds (size: $64 \times 3$), and assess the success rates of the compared methods when provided with these noisy inputs.
Fig.~\ref{fig:degraded_level} demonstrates that \mname{} exhibits markedly superior robustness to observation degradation. As the noise level increases, the success rate of DP3 declines precipitously, whereas our method maintains consistently high performance. This resilience is attributed to the model’s ability to leverage historical action sequences, thereby compensating for the diminished spatial constraints inherent in degraded observations.

\noindent \textbf{Efficiency of Cache Sharing Mechanism}.
We evaluate the efficiency of our Cache Sharing Mechanism on a single RTX 4090D by measuring per-\texttt{AR}-step inference latency while varying the length of the Historical Data Pairs (i.e. Historical Actions \(\tilde{\textbf{A}}\) and Observations \(\textbf{O}\)). 
As shown in Fig.~\ref{fig:acceleration_level}, enabling Cache Sharing Mechanism consistently reduces latency, and  this acceleration becomes more pronounced as the Historical Data Pairs lengthens.


\section{Real-World Experiments}
\label{sec:real_experiment}
\begin{table}[t!]
    \centering
    \renewcommand{\arraystretch}{1.5}
    \caption{Quantitative results of our \mname{} and the baseline method on \textbf{real-world} tasks. Succ. is the abbreviation for Success Rate.
    }
    \label{tab:real-performance}
    \resizebox{0.85\textwidth}{!}{
    \begin{tabular}{c c c c c c c c}
        \hline
        \multirow{2}{*}{Method} & \multicolumn{3}{c}{Collecting Objects} & \multicolumn{3}{c}{Stacking Cubes} & Push T \\ \cmidrule(lr){2-4} \cmidrule(lr){5-7} \cmidrule(lr){8-8} \
        & Grasping Succ. & Placing Succ. & Succ.& Grasping Succ. & Placing Succ. & Succ. & Succ. \\ \hline
        \rowcolor{lightblue}
         CDP~(Ours) & \textbf{ 18 / 20 } & \textbf{ 13 / 18 } & \textbf{ 13 / 20 } & \textbf{ 16 / 20 } & \textbf{ 10 / 16 } & \textbf{ 10 / 20 } & \textbf{ 17 / 20 } \\ 
        DP &  \textbf{ 18 / 20} & 11 / 18 & 11 / 20 & 13 / 20 & 7 / 13 & 7 / 20 & 14 / 20 \\ \hline
    \end{tabular}
    }
\end{table}

We list the quantitative results of our real-world experiments in Table~\ref{tab:real-performance}.  For the tasks of Collecting Objects and Stacking Cubes, we report not only the overall success rate but also the individual success rates for grasping and placing. The placing success rate is measured conditional on successful grasping. As shown in the table, our model outperforms the baseline method in terms of grasping, placing and overall success rate, highlighting the effectiveness of our approach. 
Figure~\ref{fig:visualization} illustrates representative trajectories of real-world tasks generated by \mname{}. 
Consistent with simulation results, real-world experiments confirm that \mname{} attains high success rates across all tasks, even when trained with a modest number of 50 demonstrations. 


\section{Conclusion}
\label{sec:conclusion}

Our proposed \mname{} effectively counteracts the impact of observation-quality degradation—arising from sensor noise, occlusions, and hardware limitations—on reliable robotic manipulation. Built upon a causal-transformer diffusion framework, the method captures critical temporal dependencies, thereby compensating for the spatial constraints diminished by degraded observations and preserving task performance. Extensive evaluations demonstrate that our \mname{} consistently outperforms existing approaches, attesting to its robustness and effectiveness. To facilitate real-time inference, we further introduce a Cache Sharing Mechanism that reuses pre-computed key–value attention tensors, yielding substantial reductions in the computational burden of autoregressive action generation without compromising accuracy.

\textbf{Limitation.} 
This study has not addressed tasks demanding extremely long horizons. The extension of our method to scenarios requiring sustained and coherent planning over extended temporal spans remains an avenue for future inquiry.

\clearpage


\bibliography{example}  

\newpage
\appendix
\input{supp}

\end{document}

%% file: supp.tex
\section{Simulation Experiments Details}

\subsection{Experiment Setup}

\textbf{Benchmarks.} 1) \textbf{Adroit}~\cite{rajeswaran2017learning}: This benchmark employs a multi-fingered Shadow robot within the MuJoCo environment to perform highly dexterous manipulation tasks, including interactions with articulated objects and rigid bodies. It is designed to test advanced manipulation skills and coordination.
2) \textbf{Dexart}~\cite{bao2023dexart}: This dataset utilizes the Allegro robot within the SAPIEN environment to execute high-precision dexterous manipulation tasks, primarily focusing on articulated object manipulation. It emphasizes fine motor skills and adaptability in complex scenarios.
3) \textbf{MetaWorld}~\cite{yu2020meta}: This benchmark operates within the MuJoCo environment, using a robotic gripper to perform a diverse range of manipulation tasks involving both articulated and rigid objects. Tasks are categorized by difficulty levels: easy, medium, hard, and very hard. Its evaluation covers tasks from medium to very hard difficulty levels.
4) \textbf{RoboFactory}~\cite{qin2025robofactory}: This is a benchmark for embodied multi-agent systems, focusing on collaborative tasks with compositional constraints to ensure safe and efficient interactions. The task challenges include designing effective coordination mechanisms to enable agents to work together seamlessly, and ensuring robustness against uncertainties and dynamic changes in the environment during long-term task execution.

\textbf{Baselines.} To systematically evaluate the performance of our proposed method \mname{}, we selected Diffusion Policy and 3D Diffusion Policy as the 2D and 3D baseline methods, respectively. Different from previous works, we employed an MLP as the visual encoder for both the 2D and 3D baseline methods in our experiments to ensure consistency. All methods were subjected to the same number of observation and inference steps, and were trained using an identical set of expert demonstrations as well as an equivalent number of training epochs.

\textbf{Evaluation Metric.} Following the established metrics~\cite{Ze2024DP3}, each experiment was conducted across three independent trials, utilizing seed values of 0, 1, and 2, in the Adroit, DexArt, and MetaWorld benchmarks. For each seed, the policy was evaluated over 20 episodes every 200 training epochs, and the mean of the top 5 success rates was computed. The final reported performance consists of the mean and standard deviation of these success rates across above three seeds.
In the Robofactory benchmark, each experiment was executed using only one seed value (\textit{i.e.} 0), and was evaluated over 100 episodes at the 300th training epoch. This comprehensive evaluation metrics ensures a fair comparison between our \mname{} and baseline methods, providing a clear assessment of the performance improvements achieved by our approach.

\subsection{Settings}
In this part, we provide detailed descriptions of all settings in our simulation experiments, including those related to the training strategy, as well as the inputs (Observations $\textbf{O}$ and Historical Actions \(\tilde{\textbf{A}}\)) and outputs (Target Actions \(\textbf{A}\)).
\begin{table}[ht!]
    \centering
    \caption{ \textbf{All Settings in Simulation Experiments.} $\mathcal{N}(0,1)$ denotes Gaussian noise characterized by a mean of 0 and a variance of 1. Following~\cite{chi2023diffusion}, the Target Actions \(\textbf{A}\) is composed of \textbf{Valid Parts} and \textbf{Redundant Parts}.}
    \resizebox{0.52\textwidth}{!}{
    \begin{tabular}{@{}lc@{}} 
        \toprule
        Parameter & Value \\ \midrule
        Length of Historical Actions & 20 \\
        Length of Target Actions (Valid Parts) & 8 \\
        Length of Target Actions (Redundant Parts) & 4 \\
        Noise in Perturbed Historical Actions & $\frac{1}{6} \times\mathcal{N}(0,1)$ \\
        Size of Point Clouds & (512, 3) \\
        Size of Images &  (84, 84, 3) \\
        Batch Size & 64 \\
        Epoch & 3000 \\
        \bottomrule
    \end{tabular}
    }
    \label{tab:simulation-parameters}
\end{table}

Specifically, we made the following changes in the simulation experiments conducted on the RoboFactory benchmark,
\begin{itemize}
\item In the 2D scenario, images were resized to (64, 64, 3) instead of (84, 84, 3). And in the 3D scenario, the size of point clouds was reduced to (128, 3), deviating from the standard configuration.

\item Noise in Perturbed Historical Actions was increased from $\frac{1}{6} \times \mathcal{N}(0,1)$ to $\frac{1}{2} \times \mathcal{N}(0,1)$. 

\item The training epoch is set to 300 in RoboFactory benchmark.
\end{itemize}

\subsection{Ablation Study}
In this section, we conduct experiments on the \textbf{Lift Barrier} task from RoboFactory~\cite{qin2025robofactory} to systematically examine the impact of individual hyperparameters on the performance of our \mname{}.

\noindent \textbf{Noise Scale in Perturbed Historical Actions}. 
To justify our design choice on noise scale in Perturbed Historical Actions, we list the quantitative results og our \mname{} on different noise scales in Table~\ref{tab:noise-scale}.
A small noise scale reduces robustness to accumulating prediction errors, whereas an excessive magnitude distorts the underlying structure of the Historical Actions and impedes convergence.
Empirically, an intermediate value of 1/6 strikes an effective balance, yielding the highest success rate across all tasks.

\begin{table}[ht!]
    \centering
    \renewcommand{\arraystretch}{1.5}
    \caption{Quantitative results of our proposed method on different noise scales injected in Perturbed Historical Actions.
    }
    \label{tab:noise-scale}
    \resizebox{0.5\linewidth}{!}{
    \begin{tabular}{c c c c c c c}
        \hline
        \multirow{2}{*}{Method} & \multicolumn{6}{c}{Noise Scale} \\ \cmidrule(lr){2-7} \
        & 1/9 & 1/6 & 1/3 & 1 & 3 & 6 \\ \hline
       CDP3 & 30\% & \textbf{40\%} & 36\% & 25\% & 25\% & 23\% \\
        \hline
    \end{tabular}
    }
\end{table}

\noindent \textbf{Chunk Size}. To justify our design choice on chunk size, we list the quantitative results of our \mname{} on different chunk sizes in Table~\ref{tab:chunk-size}. For simpler tasks, we recommend a Chunk Size of 8, and a larger size for more complex tasks.

\begin{table}[ht!]
    \centering
    \renewcommand{\arraystretch}{1.5}
    \caption{Quantitative results of our proposed method on different chunk sizes.
    }
    \label{tab:chunk-size}
    \resizebox{0.35\linewidth}{!}{
    \begin{tabular}{c c c c c}
        \hline
        \multirow{2}{*}{Method} & \multicolumn{4}{c}{Chunk Size} \\ \cmidrule(lr){2-5} \
        & 1 & 2 & 4 & 8 \\ \hline
       CDP3 & 1\% & 5\% & 35\% & \textbf{40\%} \\
        \hline
    \end{tabular}
    }
\end{table}

\noindent \textbf{Length of the Historical Actions}. The first two columns of Table~\ref{tab:history-target-performance} presents the success rates of the model for different configurations of Historical Actions length. From this table, we find that a longer sequence of Historical Actions tends to be beneficial in enhancing model performance.

\begin{table}[ht!]
    \centering
    \renewcommand{\arraystretch}{1.5}
    \caption{Quantitative results of our proposed methods across different length of Historical Actions. The format ``$*_1$ - $*_2$ - $*_3$" is used to denote the following: $*_1$ represents the length of Historical Actions. $*_2$ and $*_3$  represent the length of valid and redundant parts of Target Actions, respectively.
    }
    \label{tab:history-target-performance}
    \resizebox{0.65\linewidth}{!}{
    \begin{tabular}{c c c c c c c c c c}
        \hline
        \multirow{2}{*}{Method} & \multicolumn{3}{c}{? - 4 - 4} & \multicolumn{3}{c}{? - 4 - 8} & \multicolumn{3}{c}{8 - ? - 8} \\ \cmidrule(lr){2-4} \cmidrule(lr){5-7} \cmidrule(lr){8-10} \
        & 4 & 8 & 16 & 4 & 8 & 16 & 2 & 4 & 8 \\ \hline
        CDP3 & 0\% & 0\% & 36\% & 0\% & 6\% & 37\% & 0\% & 6\% & \textbf{33\%} \\ 
        CDP & 0\% & 0\% & \textbf{37\%} & 0 & 0\% & \textbf{39\%} & 0\% & 0\% & \textbf{38\%} \\
        \hline
    \end{tabular}
    }
\end{table}

\noindent \textbf{Length of the Denoising Targets}. The last column in Table~\ref{tab:history-target-performance} illustrates that longer Denoising Targets will provide higher gains for the model. For the whole table, we observe that the optimal length of Denoising Targets should be determined in conjunction with the length of Historical Actions. Only well-matched combinations of these action lengths can effectively enhance model performance.

\subsection{ Qualitative Results of Simulation Experiments }
\begin{figure*}[ht!]
    \centering
    \setlength{\fboxrule}{0pt}
    \framebox{{\includegraphics[width=0.9\linewidth]{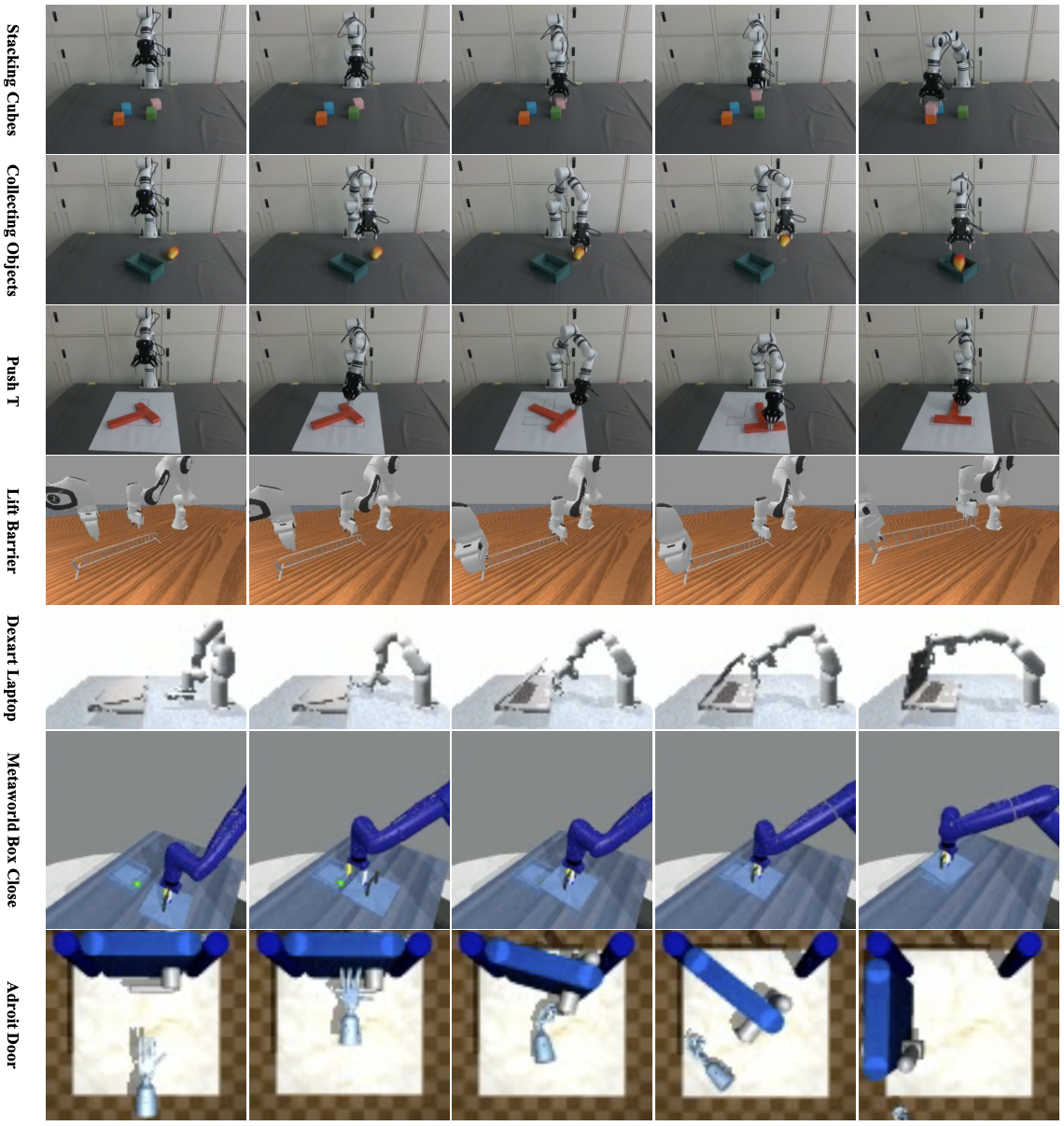}}}
    \caption{ \textbf{Visualization of real-world and simulation experimental results,} including Stacking Cubes, Collecting Objects, Push T, Lift Barrier, Dexart Laptop, Metaworld Box Close and Adroit Door task from the top to the bottom. }
    \label{fig:visualization}
\end{figure*}

Fig.~\ref{fig:visualization} visualize the results of our \mname{} across various tasks on the Adroit, Dexart, MetaWorld, and RoboFactory benchmarks, which demonstrate the effectiveness of our model in multiple simulation environments.

\section{Real-world Experiments Details}
\subsection{Workspace} 
\begin{figure*}[ht!]
    \centering
    \setlength{\fboxrule}{0pt}
    \framebox{{\includegraphics[width=0.8\linewidth]{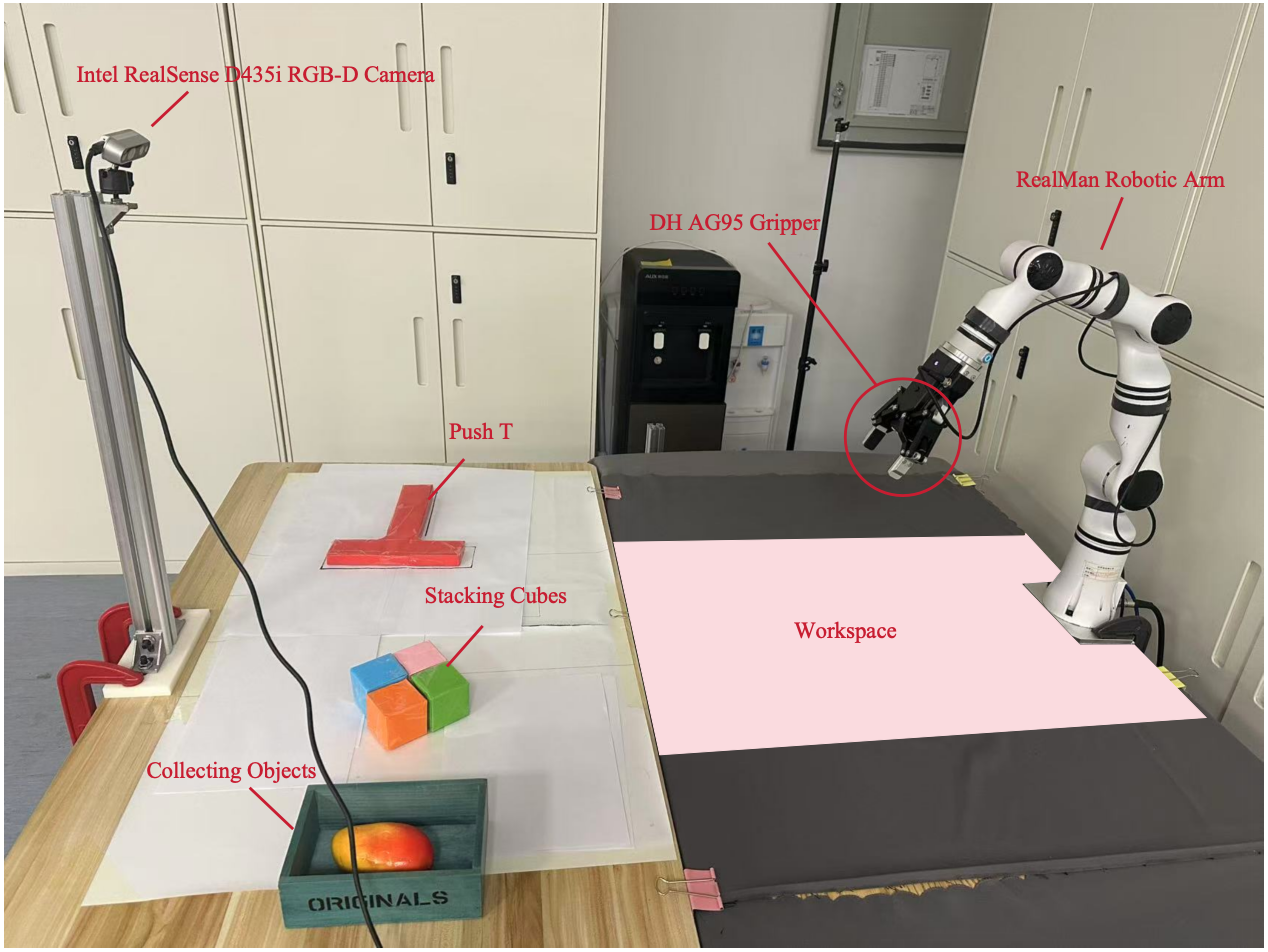}}}
    \caption{Real-world Experimental Workspace}
    \label{fig:workspace}
\end{figure*}

As illustrated in Fig.~\ref{fig:workspace}, our real-world experiments were conducted using a RealMan robotic arm fitted with a DH AG95 gripper. A stationary top-down Intel RealSense D435i RGB-D camera was employed to provide a global RGB image of the workspace. All hardware components were connected to a workstation equipped with an NVIDIA 3090 Ti GPU. This workspace enabled efficient data acquisition and evaluation of the robotic system's performance.

\subsection{Experiment Setup}

\textbf{Demonstrations.} 
The demonstrations utilized in our real-world experiments were meticulously collected via human teleoperation, facilitated by advanced vision-based retargeting techniques. For each task, we collected a total of 50 demonstrations, each carefully selected to encapsulate the fundamental skills and critical interactions necessary for successful task completion. This curation keeps the dataset compact while ensuring it faithfully represents the inherent complexities and challenges of each task.

\textbf{Settings.}
All settings in our real-world experiments are list in Table~\ref{tab:real-parameters}. Similar with our simulation experiments, these settings involve the training strategy, as well as the inputs (Observations $\textbf{O}$ and Historical Actions \(\tilde{\textbf{A}}\)) and outputs (Target Actions \(\textbf{A}\)).

\begin{table}[ht!]
    \centering
    \caption{All Settings in Real-World Experiments.}
    \resizebox{0.5\textwidth}{!}{
    \begin{tabular}{@{}lc@{}} 
        \toprule
        Parameter & Value \\ \midrule
        Length of Historical Actions & 8 \\
        Length of Target Actions (Valid Parts) & 16 \\
        Length of Target Actions (Redundant Parts) & 8 \\
        Noise in Perturbed Historical Actions & $\frac{1}{2} \times\mathcal{N}(0,1)$ \\
        Batch Size & 64 \\
        Size of Images &  (60, 80, 3) \\
        Epoch & 300 \\
        \bottomrule
    \end{tabular}
    }
    \label{tab:real-parameters}
\end{table}

\subsection{Tasks}
In real-world scenarios, we evaluate our proposed model and the baseline methods through three tasks: Collecting Objects, Stacking Cubes, and Push-T. The completion criteria for each task are described as follows:
\begin{itemize}
\item \textbf{Collecting Objects} is considered successful if the robot places the mango into the green box. This task not only assesses the ability to grasp the object but also the precision in placing it accurately into the designated green box. 
\item \textbf{Stacking Cubes} involves four colored cubes: blue, green, pink, and orange. This task is successful if the robot stacks the pink cube on top of the orange one while ignoring the blue and green cubes, which serve as distractors.
\item \textbf{Push T}: The robot must rotate and translate a T-shaped block to position it within a specified range to achieve task success.
\end{itemize}


\clearpage




%% file: example.bbl
\begin{thebibliography}{49}
\providecommand{\natexlab}[1]{#1}
\providecommand{\url}[1]{\texttt{#1}}
\expandafter\ifx\csname urlstyle\endcsname\relax
  \providecommand{\doi}[1]{doi: #1}\else
  \providecommand{\doi}{doi: \begingroup \urlstyle{rm}\Url}\fi

\bibitem[Shridhar et~al.(2023)Shridhar, Manuelli, and Fox]{shridhar2023perceiver}
M.~Shridhar, L.~Manuelli, and D.~Fox.
\newblock Perceiver-actor: A multi-task transformer for robotic manipulation.
\newblock In \emph{Conference on Robot Learning}, pages 785--799. PMLR, 2023.

\bibitem[Wang et~al.(2023)Wang, Fan, Sun, Zhang, Fei-Fei, Xu, Zhu, and Anandkumar]{wang2023mimicplay}
C.~Wang, L.~Fan, J.~Sun, R.~Zhang, L.~Fei-Fei, D.~Xu, Y.~Zhu, and A.~Anandkumar.
\newblock Mimicplay: Long-horizon imitation learning by watching human play.
\newblock \emph{arXiv preprint arXiv:2302.12422}, 2023.

\bibitem[Ze et~al.(2023)Ze, Yan, Wu, Macaluso, Ge, Ye, Hansen, Li, and Wang]{ze2023gnfactor}
Y.~Ze, G.~Yan, Y.-H. Wu, A.~Macaluso, Y.~Ge, J.~Ye, N.~Hansen, L.~E. Li, and X.~Wang.
\newblock Gnfactor: Multi-task real robot learning with generalizable neural feature fields.
\newblock In \emph{Conference on Robot Learning}, pages 284--301. PMLR, 2023.

\bibitem[Peng et~al.(2020)Peng, Coumans, Zhang, Lee, Tan, and Levine]{peng2020learning}
X.~B. Peng, E.~Coumans, T.~Zhang, T.-W. Lee, J.~Tan, and S.~Levine.
\newblock Learning agile robotic locomotion skills by imitating animals.
\newblock \emph{arXiv preprint arXiv:2004.00784}, 2020.

\bibitem[Agarwal et~al.(2023)Agarwal, Uppal, Shaw, and Pathak]{agarwal2023dexterous}
A.~Agarwal, S.~Uppal, K.~Shaw, and D.~Pathak.
\newblock Dexterous functional grasping.
\newblock \emph{arXiv preprint arXiv:2312.02975}, 2023.

\bibitem[Haldar et~al.(2023)Haldar, Pari, Rai, and Pinto]{haldar2023teach}
S.~Haldar, J.~Pari, A.~Rai, and L.~Pinto.
\newblock Teach a robot to fish: Versatile imitation from one minute of demonstrations.
\newblock \emph{arXiv preprint arXiv:2303.01497}, 2023.

\bibitem[Chi et~al.(2023)Chi, Feng, Du, Xu, Cousineau, Burchfiel, and Song]{chi2023diffusion}
C.~Chi, S.~Feng, Y.~Du, Z.~Xu, E.~Cousineau, B.~Burchfiel, and S.~Song.
\newblock Diffusion policy: Visuomotor policy learning via action diffusion.
\newblock \emph{arXiv preprint arXiv:2303.04137}, 2023.

\bibitem[Ho et~al.(2020)Ho, Jain, and Abbeel]{ho2020denoising}
J.~Ho, A.~Jain, and P.~Abbeel.
\newblock Denoising diffusion probabilistic models.
\newblock \emph{Advances in neural information processing systems}, 33:\penalty0 6840--6851, 2020.

\bibitem[Song et~al.(2020)Song, Meng, and Ermon]{song2020denoising}
J.~Song, C.~Meng, and S.~Ermon.
\newblock Denoising diffusion implicit models.
\newblock \emph{arXiv preprint arXiv:2010.02502}, 2020.

\bibitem[Janner et~al.(2022)Janner, Du, Tenenbaum, and Levine]{janner2022planning}
M.~Janner, Y.~Du, J.~B. Tenenbaum, and S.~Levine.
\newblock Planning with diffusion for flexible behavior synthesis.
\newblock \emph{arXiv preprint arXiv:2205.09991}, 2022.

\bibitem[Luo et~al.(2024)Luo, Sun, Tenenbaum, and Du]{luo2024potential}
Y.~Luo, C.~Sun, J.~B. Tenenbaum, and Y.~Du.
\newblock Potential based diffusion motion planning.
\newblock \emph{arXiv preprint arXiv:2407.06169}, 2024.

\bibitem[Carvalho et~al.(2023)Carvalho, Le, Baierl, Koert, and Peters]{carvalho2023motion}
J.~Carvalho, A.~T. Le, M.~Baierl, D.~Koert, and J.~Peters.
\newblock Motion planning diffusion: Learning and planning of robot motions with diffusion models.
\newblock In \emph{2023 IEEE/RSJ International Conference on Intelligent Robots and Systems (IROS)}, pages 1916--1923. IEEE, 2023.

\bibitem[Saha et~al.(2024)Saha, Mandadi, Reddy, Srikanth, Agarwal, Sen, Singh, and Krishna]{saha2024edmp}
K.~Saha, V.~Mandadi, J.~Reddy, A.~Srikanth, A.~Agarwal, B.~Sen, A.~Singh, and M.~Krishna.
\newblock Edmp: Ensemble-of-costs-guided diffusion for motion planning.
\newblock In \emph{2024 IEEE International Conference on Robotics and Automation (ICRA)}, pages 10351--10358. IEEE, 2024.

\bibitem[Zhou et~al.(2024)Zhou, Qin, Yin, Huang, Zhang, Sheng, Qiao, and Shao]{zhou2024minedreamer}
E.~Zhou, Y.~Qin, Z.~Yin, Y.~Huang, R.~Zhang, L.~Sheng, Y.~Qiao, and J.~Shao.
\newblock Minedreamer: Learning to follow instructions via chain-of-imagination for simulated-world control.
\newblock \emph{arXiv preprint arXiv:2403.12037}, 2024.

\bibitem[Qin et~al.(2025)Qin, Sun, Hong, Wang, and Zhang]{qin2025navigatediff}
Y.~Qin, A.~Sun, Y.~Hong, B.~Wang, and R.~Zhang.
\newblock Navigatediff: Visual predictors are zero-shot navigation assistants.
\newblock \emph{arXiv preprint arXiv:2502.13894}, 2025.

\bibitem[Huang et~al.(2023)Huang, Wang, Li, Jia, Liu, Zhu, Liang, and Zhu]{huang2023diffusion}
S.~Huang, Z.~Wang, P.~Li, B.~Jia, T.~Liu, Y.~Zhu, W.~Liang, and S.-C. Zhu.
\newblock Diffusion-based generation, optimization, and planning in 3d scenes.
\newblock In \emph{Proceedings of the IEEE/CVF Conference on Computer Vision and Pattern Recognition}, pages 16750--16761, 2023.

\bibitem[Pearce et~al.(2023)Pearce, Rashid, Kanervisto, Bignell, Sun, Georgescu, Macua, Tan, Momennejad, Hofmann, et~al.]{pearce2023imitating}
T.~Pearce, T.~Rashid, A.~Kanervisto, D.~Bignell, M.~Sun, R.~Georgescu, S.~V. Macua, S.~Z. Tan, I.~Momennejad, K.~Hofmann, et~al.
\newblock Imitating human behaviour with diffusion models.
\newblock \emph{arXiv preprint arXiv:2301.10677}, 2023.

\bibitem[Ha et~al.(2023)Ha, Florence, and Song]{ha2023scaling}
H.~Ha, P.~Florence, and S.~Song.
\newblock Scaling up and distilling down: Language-guided robot skill acquisition.
\newblock In \emph{Conference on Robot Learning}, pages 3766--3777. PMLR, 2023.

\bibitem[Xian et~al.(2023)Xian, Gkanatsios, Gervet, Ke, and Fragkiadaki]{xian2023chaineddiffuser}
Z.~Xian, N.~Gkanatsios, T.~Gervet, T.-W. Ke, and K.~Fragkiadaki.
\newblock Chaineddiffuser: Unifying trajectory diffusion and keypose prediction for robotic manipulation.
\newblock In \emph{7th Annual Conference on Robot Learning}, 2023.

\bibitem[Li et~al.(2024)Li, Belagali, Shang, and Ryoo]{li2024crossway}
X.~Li, V.~Belagali, J.~Shang, and M.~S. Ryoo.
\newblock Crossway diffusion: Improving diffusion-based visuomotor policy via self-supervised learning.
\newblock In \emph{2024 IEEE International Conference on Robotics and Automation (ICRA)}, pages 16841--16849. IEEE, 2024.

\bibitem[Wang et~al.(2024)Wang, Zhao, Du, Adelson, and Tedrake]{wang2024poco}
L.~Wang, J.~Zhao, Y.~Du, E.~H. Adelson, and R.~Tedrake.
\newblock Poco: Policy composition from and for heterogeneous robot learning.
\newblock \emph{arXiv preprint arXiv:2402.02511}, 2024.

\bibitem[Chen et~al.(2024)Chen, Lim, Lin, Chen, and Soh]{chen2024don}
K.~Chen, E.~Lim, K.~Lin, Y.~Chen, and H.~Soh.
\newblock Don't start from scratch: Behavioral refinement via interpolant-based policy diffusion.
\newblock \emph{arXiv preprint arXiv:2402.16075}, 2024.

\bibitem[Sridhar et~al.(2023)Sridhar, Dutta, Jayaraman, Weimer, and Lee]{sridhar2023memory}
K.~Sridhar, S.~Dutta, D.~Jayaraman, J.~Weimer, and I.~Lee.
\newblock Memory-consistent neural networks for imitation learning.
\newblock \emph{arXiv preprint arXiv:2310.06171}, 2023.

\bibitem[Zhao et~al.(2024)Zhao, Tompson, Driess, Florence, Ghasemipour, Finn, and Wahid]{zhao2024aloha}
T.~Z. Zhao, J.~Tompson, D.~Driess, P.~Florence, K.~Ghasemipour, C.~Finn, and A.~Wahid.
\newblock Aloha unleashed: A simple recipe for robot dexterity.
\newblock \emph{arXiv preprint arXiv:2410.13126}, 2024.

\bibitem[Chi et~al.(2024)Chi, Xu, Pan, Cousineau, Burchfiel, Feng, Tedrake, and Song]{chi2024universal}
C.~Chi, Z.~Xu, C.~Pan, E.~Cousineau, B.~Burchfiel, S.~Feng, R.~Tedrake, and S.~Song.
\newblock Universal manipulation interface: In-the-wild robot teaching without in-the-wild robots.
\newblock \emph{arXiv preprint arXiv:2402.10329}, 2024.

\bibitem[Kang et~al.(2025)Kang, Song, Zhou, Qin, Yang, Liu, Torr, Bai, and Yin]{kang2025viki}
L.~Kang, X.~Song, H.~Zhou, Y.~Qin, J.~Yang, X.~Liu, P.~Torr, L.~Bai, and Z.~Yin.
\newblock Viki-r: Coordinating embodied multi-agent cooperation via reinforcement learning.
\newblock \emph{arXiv preprint arXiv:2506.09049}, 2025.

\bibitem[Ze et~al.(2024)Ze, Zhang, Zhang, Hu, Wang, and Xu]{Ze2024DP3}
Y.~Ze, G.~Zhang, K.~Zhang, C.~Hu, M.~Wang, and H.~Xu.
\newblock 3d diffusion policy: Generalizable visuomotor policy learning via simple 3d representations.
\newblock In \emph{Proceedings of Robotics: Science and Systems (RSS)}, 2024.

\bibitem[Fu et~al.(2024)Fu, Huang, Datta, Chen, Panitch, Liu, Li, and Goldberg]{fu2024context}
L.~Fu, H.~Huang, G.~Datta, L.~Y. Chen, W.~C.-H. Panitch, F.~Liu, H.~Li, and K.~Goldberg.
\newblock In-context imitation learning via next-token prediction.
\newblock \emph{arXiv preprint arXiv:2408.15980}, 2024.

\bibitem[Zhang et~al.(2025)Zhang, Liu, Chang, Schramm, and Boularias]{zhang2025autoregressive}
X.~Zhang, Y.~Liu, H.~Chang, L.~Schramm, and A.~Boularias.
\newblock Autoregressive action sequence learning for robotic manipulation.
\newblock \emph{IEEE Robotics and Automation Letters}, 2025.

\bibitem[Gong et~al.(2024)Gong, Ding, Lyu, Huang, Sun, Zhao, Fan, and Wang]{gong2024carp}
Z.~Gong, P.~Ding, S.~Lyu, S.~Huang, M.~Sun, W.~Zhao, Z.~Fan, and D.~Wang.
\newblock Carp: Visuomotor policy learning via coarse-to-fine autoregressive prediction.
\newblock \emph{arXiv preprint arXiv:2412.06782}, 2024.

\bibitem[Chen et~al.(2023)Chen, Yu, Ge, Yao, Xie, Wu, Wang, Kwok, Luo, Lu, et~al.]{chen2023pixart}
J.~Chen, J.~Yu, C.~Ge, L.~Yao, E.~Xie, Y.~Wu, Z.~Wang, J.~Kwok, P.~Luo, H.~Lu, et~al.
\newblock Pixart-$\alpha $: Fast training of diffusion transformer for photorealistic text-to-image synthesis.
\newblock \emph{arXiv preprint arXiv:2310.00426}, 2023.

\bibitem[Peebles and Xie(2023)]{peebles2023scalable}
W.~Peebles and S.~Xie.
\newblock Scalable diffusion models with transformers.
\newblock In \emph{Proceedings of the IEEE/CVF international conference on computer vision}, pages 4195--4205, 2023.

\bibitem[Qin et~al.(2024)Qin, Shi, Yu, Wang, Zhou, Li, Yin, Liu, Sheng, Shao, et~al.]{qin2024worldsimbench}
Y.~Qin, Z.~Shi, J.~Yu, X.~Wang, E.~Zhou, L.~Li, Z.~Yin, X.~Liu, L.~Sheng, J.~Shao, et~al.
\newblock Worldsimbench: Towards video generation models as world simulators.
\newblock \emph{arXiv preprint arXiv:2410.18072}, 2024.

\bibitem[Rombach et~al.(2022)Rombach, Blattmann, Lorenz, Esser, and Ommer]{rombach2022high}
R.~Rombach, A.~Blattmann, D.~Lorenz, P.~Esser, and B.~Ommer.
\newblock High-resolution image synthesis with latent diffusion models.
\newblock In \emph{Proceedings of the IEEE/CVF conference on computer vision and pattern recognition}, pages 10684--10695, 2022.

\bibitem[Chen et~al.(2024)Chen, Mart{\'\i}~Mons{\'o}, Du, Simchowitz, Tedrake, and Sitzmann]{chen2024diffusion}
B.~Chen, D.~Mart{\'\i}~Mons{\'o}, Y.~Du, M.~Simchowitz, R.~Tedrake, and V.~Sitzmann.
\newblock Diffusion forcing: Next-token prediction meets full-sequence diffusion.
\newblock \emph{Advances in Neural Information Processing Systems}, 37:\penalty0 24081--24125, 2024.

\bibitem[Yin et~al.(2024)Yin, Zhang, Zhang, Freeman, Durand, Shechtman, and Huang]{yin2024slow}
T.~Yin, Q.~Zhang, R.~Zhang, W.~T. Freeman, F.~Durand, E.~Shechtman, and X.~Huang.
\newblock From slow bidirectional to fast causal video generators.
\newblock \emph{arXiv preprint arXiv:2412.07772}, 2024.

\bibitem[Guo et~al.(2023)Guo, Yang, Rao, Liang, Wang, Qiao, Agrawala, Lin, and Dai]{guo2023animatediff}
Y.~Guo, C.~Yang, A.~Rao, Z.~Liang, Y.~Wang, Y.~Qiao, M.~Agrawala, D.~Lin, and B.~Dai.
\newblock Animatediff: Animate your personalized text-to-image diffusion models without specific tuning.
\newblock \emph{arXiv preprint arXiv:2307.04725}, 2023.

\bibitem[Lu et~al.(2023)Lu, Yang, Fei, Huo, Lu, Luo, and Ding]{lu2023vdt}
H.~Lu, G.~Yang, N.~Fei, Y.~Huo, Z.~Lu, P.~Luo, and M.~Ding.
\newblock Vdt: General-purpose video diffusion transformers via mask modeling.
\newblock \emph{arXiv preprint arXiv:2305.13311}, 2023.

\bibitem[Ma et~al.(2024)Ma, Wang, Jia, Chen, Liu, Li, Chen, and Qiao]{ma2024latte}
X.~Ma, Y.~Wang, G.~Jia, X.~Chen, Z.~Liu, Y.-F. Li, C.~Chen, and Y.~Qiao.
\newblock Latte: Latent diffusion transformer for video generation.
\newblock \emph{arXiv preprint arXiv:2401.03048}, 2024.

\bibitem[Ren et~al.(2024)Ren, Yang, Zhang, Wei, Du, Huang, and Chen]{ren2024consisti2v}
W.~Ren, H.~Yang, G.~Zhang, C.~Wei, X.~Du, W.~Huang, and W.~Chen.
\newblock Consisti2v: Enhancing visual consistency for image-to-video generation.
\newblock \emph{arXiv preprint arXiv:2402.04324}, 2024.

\bibitem[Yu et~al.(2025{\natexlab{a}})Yu, Qin, Che, Liu, Wang, Wan, Zhang, and Liu]{yu2025position}
J.~Yu, Y.~Qin, H.~Che, Q.~Liu, X.~Wang, P.~Wan, D.~Zhang, and X.~Liu.
\newblock Position: Interactive generative video as next-generation game engine.
\newblock \emph{arXiv preprint arXiv:2503.17359}, 2025{\natexlab{a}}.

\bibitem[Yu et~al.(2025{\natexlab{b}})Yu, Qin, Che, Liu, Wang, Wan, Zhang, Gai, Chen, and Liu]{yu2025survey}
J.~Yu, Y.~Qin, H.~Che, Q.~Liu, X.~Wang, P.~Wan, D.~Zhang, K.~Gai, H.~Chen, and X.~Liu.
\newblock A survey of interactive generative video.
\newblock \emph{arXiv preprint arXiv:2504.21853}, 2025{\natexlab{b}}.

\bibitem[Yu et~al.(2025{\natexlab{c}})Yu, Qin, Wang, Wan, Zhang, and Liu]{yu2025gamefactory}
J.~Yu, Y.~Qin, X.~Wang, P.~Wan, D.~Zhang, and X.~Liu.
\newblock Gamefactory: Creating new games with generative interactive videos.
\newblock \emph{arXiv preprint arXiv:2501.08325}, 2025{\natexlab{c}}.

\bibitem[Yu et~al.(2025{\natexlab{d}})Yu, Bai, Qin, Liu, Wang, Wan, Zhang, and Liu]{yu2025cam}
J.~Yu, J.~Bai, Y.~Qin, Q.~Liu, X.~Wang, P.~Wan, D.~Zhang, and X.~Liu.
\newblock Context as memory: Scene-consistent interactive long video generation with memory retrieval.
\newblock \emph{arXiv preprint arXiv:2506.03141}, 2025{\natexlab{d}}.

\bibitem[Gao et~al.(2024)Gao, Shi, Zhang, Wang, Xiao, and Chen]{gao2024ca2}
K.~Gao, J.~Shi, H.~Zhang, C.~Wang, J.~Xiao, and L.~Chen.
\newblock Ca2-vdm: Efficient autoregressive video diffusion model with causal generation and cache sharing.
\newblock \emph{arXiv preprint arXiv:2411.16375}, 2024.

\bibitem[Qin et~al.(2025)Qin, Kang, Song, Yin, Liu, Liu, Zhang, and Bai]{qin2025robofactory}
Y.~Qin, L.~Kang, X.~Song, Z.~Yin, X.~Liu, X.~Liu, R.~Zhang, and L.~Bai.
\newblock Robofactory: Exploring embodied agent collaboration with compositional constraints.
\newblock \emph{arXiv preprint arXiv:2503.16408}, 2025.

\bibitem[Rajeswaran et~al.(2017)Rajeswaran, Kumar, Gupta, Vezzani, Schulman, Todorov, and Levine]{rajeswaran2017learning}
A.~Rajeswaran, V.~Kumar, A.~Gupta, G.~Vezzani, J.~Schulman, E.~Todorov, and S.~Levine.
\newblock Learning complex dexterous manipulation with deep reinforcement learning and demonstrations.
\newblock \emph{arXiv preprint arXiv:1709.10087}, 2017.

\bibitem[Bao et~al.(2023)Bao, Xu, Qin, and Wang]{bao2023dexart}
C.~Bao, H.~Xu, Y.~Qin, and X.~Wang.
\newblock Dexart: Benchmarking generalizable dexterous manipulation with articulated objects.
\newblock In \emph{Proceedings of the IEEE/CVF Conference on Computer Vision and Pattern Recognition}, pages 21190--21200, 2023.

\bibitem[Yu et~al.(2020)Yu, Quillen, He, Julian, Hausman, Finn, and Levine]{yu2020meta}
T.~Yu, D.~Quillen, Z.~He, R.~Julian, K.~Hausman, C.~Finn, and S.~Levine.
\newblock Meta-world: A benchmark and evaluation for multi-task and meta reinforcement learning.
\newblock In \emph{Conference on robot learning}, pages 1094--1100. PMLR, 2020.

\end{thebibliography}
